\pdfoutput=1

\documentclass[11pt]{article}

\usepackage{acl}

\usepackage{times}
\usepackage{latexsym}

\usepackage[T1]{fontenc}

\usepackage[utf8]{inputenc}

\usepackage{microtype}

\usepackage{graphicx}
\usepackage{subfig}
\usepackage{enumitem}

\usepackage{hyperref}
\usepackage{graphicx}
\usepackage{bm}
\usepackage{amsmath}
\usepackage{amsfonts}
\usepackage{color}
\usepackage{multirow}
\usepackage{booktabs}
\newcommand{\tabincell}[2]{\begin{tabular}{@{}#1@{}}#2\end{tabular}}

\newcommand{\lx}[1]{\textcolor{black}{{#1}}}

\newcommand{\lxCR}[1]{\textcolor{black}{{#1}}}

\usepackage{array}

\makeatletter
\newcommand{\thickhline}{%
    \noalign {\ifnum 0=`}\fi \hrule height 2pt
    \futurelet \reserved@a \@xhline
}
\newcolumntype{"}{@{\hskip\tabcolsep\vrule width 2pt\hskip\tabcolsep}}
\makeatother

\newcommand\mypound{\scalebox{1.2}{\raisebox{-0.1ex}{\#}}}

\title{Disentangled Learning of Stance and Aspect Topics for Vaccine Attitude Detection in Social Media}

\author{
 Lixing Zhu\textsuperscript{1}, Zheng Fang\textsuperscript{1}, Gabriele Pergola\textsuperscript{1}, Rob Procter\textsuperscript{1,2}, Yulan He\textsuperscript{1,2}\\
 \textsuperscript{1}Department of Computer Science, University of Warwick, UK\\
 \textsuperscript{2}The Alan Turing Institute, UK\\
 {\tt \{lixing.zhu,z.fang.4,gabriele.pergola.1}\\
 {\tt rob.procter,yulan.he\}@warwick.ac.uk} \\
}

\begin{document}
\maketitle
\begin{abstract}
Building models to detect vaccine attitudes on social media is challenging because of the composite, often intricate aspects involved, and the limited availability of annotated data.
Existing approaches have relied heavily on supervised training that requires abundant annotations and pre-defined aspect categories. Instead, with the aim of leveraging the large amount of unannotated data now available on vaccination, we propose a novel semi-supervised approach for vaccine attitude detection, called \textsc{VADet}.
A variational autoencoding architecture based on language models is employed to learn from unlabelled data the topical information of the domain. Then, the model is fine-tuned with a few manually annotated examples of user attitudes. We validate the effectiveness of \textsc{VADet} on our annotated data and also on an existing vaccination corpus annotated with opinions on vaccines. Our results show that \textsc{VADet} is able to learn disentangled stance and aspect topics, and outperforms existing aspect-based sentiment analysis models on both stance detection and tweet clustering. Our source code and dataset are available at \url{http://github.com/somethingx1202/VADet}.


\end{abstract}

\section{Introduction}

The aim of vaccine attitude detection in social media is to extract people's opinions towards vaccines by analysing their  online posts. This is closely related to aspect-based sentiment analysis in which both aspects and related sentiments need to be identified. Previous research has been largely focused on product reviews and relied on aspect-level sentiment annotations to train models~\cite{barnes-etal-2021-structured}, where aspect-opinions are extracted as triples~\cite{Peng_Xu_Bing_Huang_Lu_Si_2020}, polarized targets~\cite{AAAI1816541} or sentiment spans~\cite{he-etal-2019-interactive}. However, for the task of vaccine attitude detection on Twitter, such a volume of annotated data is barely available~\cite{kunneman2020monitoring,PAUL2021100012}. This scarcity of data is compounded by the diversity of attitudes, making it difficult for models to identify all aspects discussed in posts~\cite{morante-etal-2020-annotating}. 

\begin{figure}[tp]
\centering
\includegraphics[width=0.49\textwidth]{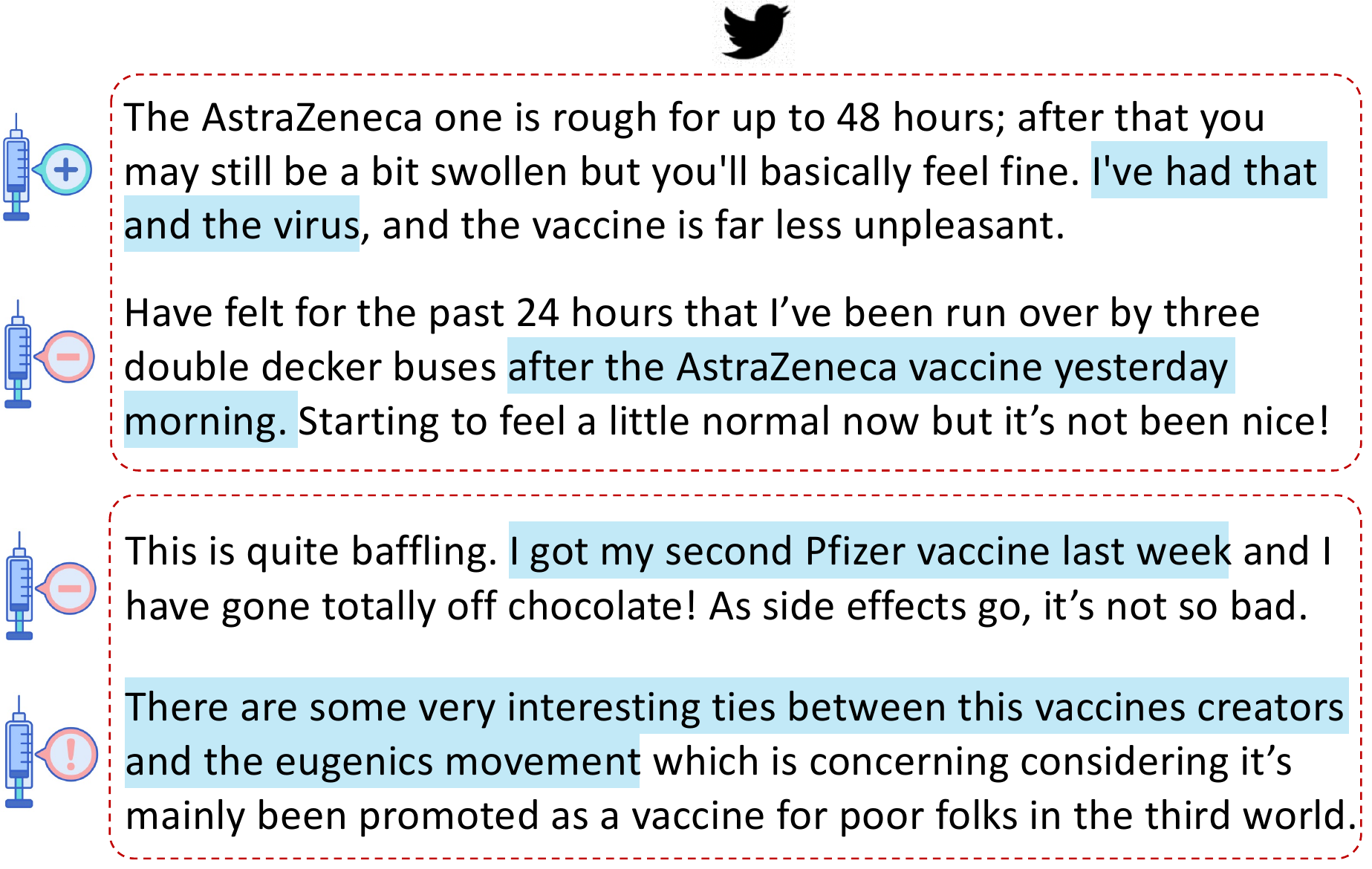}
\caption{\textbf{Top:} Expressions of aspects entangled with expressions of opinions. \textbf{Bottom:} Vaccine attitudes can be expressed towards a wide range of aspects/topics relating to vaccination, making it difficult to pre-define a set of aspect labels as opposed to corpora typically used for aspect-based sentiment analysis. } 
\vspace{-10pt}
\label{fig:0} 
\end{figure}

\lx{As representative examples, consider the two tweets about personal experiences for vaccination at the top of Figure~\ref{fig:0}. The two tweets, despite addressing a common aspect (vaccine side-effects), express opposite stances towards vaccines. However, the aspect and the stances are so fused together that the whole of the tweets need to be considered to derive the proper labels, making it difficult to disentangle them using existing methodologies.
Additionally, in the case of vaccines attitude analysis, there is a wide variety of possible aspects discussed in posts, as shown in the bottom of Figure~\ref{fig:0}, where one tweet ironically addressed vaccine side-effects and the second one expressed instead specific political concerns.
This is different from traditional aspect-based sentiment analysis on product reviews where only a small number of aspects need to be pre-defined.} 

\lx{The recently developed framework for integrating Variational Auto-Encoder (VAE)~\cite{kingma2014auto} and Independent Component Analysis (ICA)~\cite{pmlr-v108-khemakhem20a} sheds light on this problem. VAE is an unsupervised method that can be used to glean information that must be retained from the vaccine-related corpus. Meanwhile, a handful of annotations would induce the separation of independent factors following the ICA requirement for prior knowledge and inductive biases~\cite{pmlr-v89-hyvarinen19a,pmlr-v119-locatello20a,locatello2020disentangling}. To this end, we could disentangle the latent factors that are either specific to the aspect or to the stance, and improve the quality of the latent semantics learned from unannotated data.}


We frame the problem of vaccine attitude detection as a joint aspect span detection and stance classification task, assuming that a tweet, which is limited to 280 characters, would usually only discuss one aspect. In particular, we extend a pre-trained language model (LM) by adding a topic layer, which aims to model the topical theme discussed in a tweet. In the absence of annotated data, the topic layer is trained to reconstruct the input message built on VAE. Given the annotated data, where each tweet is annotated with an aspect span and a stance label, the learned topic can be disentangled into a stance topic and an aspect topic. The stance topic is used to predict the stance label of the given tweet, while the aspect topic is used to predict the start and the ending positions of the aspect span. By doing so, we can effectively leverage both unannotated and annotated data for model training. 

To evaluate the effectiveness of our proposed model for vaccine attitude detection on Twitter, we have collected over 1.9 million tweets relating to COVID vaccines between \lx{February} and April 2021. We have further annotated 2,800 tweets with both aspect spans and stance labels. 
In addition, we have also used an existing Vaccination Corpus\footnote{https://github.com/cltl/VaccinationCorpus} in which 294 documents related to the online vaccination debate have been annotated with opinions towards vaccination. Our experimental results on both datasets show that the proposed model outperforms existing opinion triple extraction model and BERT QA model on both aspect span extraction and stance classification. Moreover, the learned latent aspect topics allow the clustering of user attitudes towards vaccines, facilitating easier discovery of positive and negative attitudes in social media. \lx{The contribution of this work can be summarised as follows:}
\begin{itemize}
    \item We have proposed a novel semi-supervised approach for joint latent stance/aspect representation learning and aspect span detection;
    \item The developed disentangled representation learning facilitates better attitude detection and clustering;
    \item We have constructed an annotated dataset for vaccine attitude detection.
\end{itemize}

\section{Related Work}

Our work is related to three lines of research: aspect-based sentiment analysis, disentangled representation learning, and vaccine attitude detection.

\paragraph{Aspect-Based Sentiment Analysis} 
(ABSA) aims to identify the aspect terms and their polarities from text. \lxCR{Much work has been focusing on this task. 
The techniques used include Conditional Random Fields (CRFs)~\cite{10.1007/978-3-319-06028-6_23}, Bidirectional Long Short-Term Memory networks (BiLSTMs)~\cite{baziotis-etal-2017-datastories-semeval}, Convolutional Neural Networks (CNNs)~\cite{10.5555/2969239.2969312}, Attention Networks~\cite{yang-etal-2016-hierarchical,pergola21bem}, DenseLSTMs~\cite{wu-etal-2018-thu}, NestedLSTMs~\cite{pmlr-v77-moniz17a}, Graph Neural Networks~\cite{zhang-etal-2019-aspect} and their combinations~\cite{ijcai2018-617,zhu21topic,Wan_Yang_Du_Liu_Qi_Pan_2020}, to name a few.}

\lx{\citet{zhang-etal-2015-neural} framed this task as text span detection, where they used text spans to denote aspects. The same annotation scheme was employed in~\cite{li-etal-2018-transformation}, where intra-word attentions were designed to enrich the representations of aspects and predict their polarities. \citet{ijcai2018-583} formalized the task as a sequence labeling problem under a unified tagging scheme. Their follow-up work~\cite{li-etal-2019-exploiting} explored BERT for end-to-end ABSA. \citet{Peng_Xu_Bing_Huang_Lu_Si_2020} modified this task by introducing opinion terms to shape the polarity. A similar modification was made in~\cite{zhao-etal-2020-spanmlt} to extract aspect-opinion pairs. Position-aware tagging was introduced to entrench the offset between the aspect span and opinion term~\cite{xu-etal-2020-position}. More recently, instead of using pipeline approaches or sequence tagging, \citet{barnes-etal-2021-structured} adapted syntactic dependency parsing to perform aspect and opinion expression extraction, and polarity classification, thus formalizing the task as structured sentiment analysis.}

\paragraph{Disentangled representation learning} Deep generative models learn the hidden semantics of text, 
of which many attempt to capture the independent latent factor to steer the generation of text in the context of NLP~\cite{pmlr-v70-hu17e,li-etal-2018-delete,pergola19tdam,john-etal-2019-disentangled,li-etal-2020-optimus}. The majority of the aforementioned work employs VAE~\cite{10.5555/2969033.2969226} to learn controllable factors, leading to the abundance of VAE-based models in disentangled representation learning~\cite{DBLP:conf/iclr/HigginsMPBGBML17,burgess2018understanding,NEURIPS2018_1ee3dfcd}. However, previous studies show that unsupervised learning of disentanglement by optimising the marginal likelihood in a generative model is impossible~\cite{locatello2019challenging}. While it is also the case that non-linear ICA is unable to uncover the true independent factors, \citet{pmlr-v108-khemakhem20a} established a connection between those two strands of work, which is of particular interest to us since the proposed framework learns to approximate the true factorial prior given few examples, recovering a disentangled latent variable distribution \lxCR{on top of additionally
observed variables}. \lxCR{In this paper, stance labels and aspect spans are additionally observed on a handful of data, which could be used as inductive biases that make disentanglement possible.} 



\paragraph{Vaccine attitude detection} Very little literature exists on attitude detection for vaccination. In contrast, there is growing interest in Covid-19 corpus construction~\cite{shuja2021covid}. Of particular interest to us, \citet{banda2021large} built an on-going tweet dataset that traces the development of Covid-19 by 3 keywords: ``coronavirus'', ``2019nCoV'' and ``corona virus''. \citet{info:doi/10.2196/26627} utilized hydrated tweets from the aforementioned corpus to analyze the sentiment towards vaccination. They used lexicon-based methods (i.e., VADER and TextBlob) and pre-trained BERT to classify the sentiment in order to gain insights into the temporal sentiment trends. A similar approach has been proposed in~\cite{info:doi/10.2196/30854}. \citet{lyu2021covid} employed a topic model to discover vaccine-related themes in twitter discussions and performed sentiment classification using lexicon-based methods. However, none of the work above constructed datasets about vaccine attitudes, nor did they train models to detect attitudes. \citet{morante-etal-2020-annotating} built the Vaccination Corpus (VC) with events, attributions and opinions annotated in the form of text spans, which is the only dataset available to us to perform attitude detection.




\section{Methodology}
\lx{The goal of our work is to detect the stance expressed in a tweet (i.e., `\emph{pro-vaccination}', `\emph{anti-vaccination}', or `\emph{neutral}'), identify a text span that indicates the concerning aspect of vaccination, and cluster tweets into groups that share similar aspects. To this end, we propose a novel latent representation learning model that jointly learns a stance classifier and disentangles the latent variables capturing stance and aspect respectively. Our proposed Vaccine Attitude Detection (\textsc{VADet}) model is firstly trained on a large amount of unannotated Twitter data to learn latent topics via masked Language Model (LM) learning. It is then fine-tuned on a small amount of Twitter data annotated with stance labels and aspect text spans for simultaneously stance classification and aspect span start/end position detection. 
The rationale is that 
the inductive bias imposed by the annotations would encourage the disentanglement of latent stance topics and aspect topics.}
In what follows, we will present our proposed \textsc{VADet} model, first under the masked LM learning and later extended to the supervised setting for learning disentangled stance and aspect topics.

\begin{figure}[htb]
\centering
\includegraphics[width=0.49\textwidth]{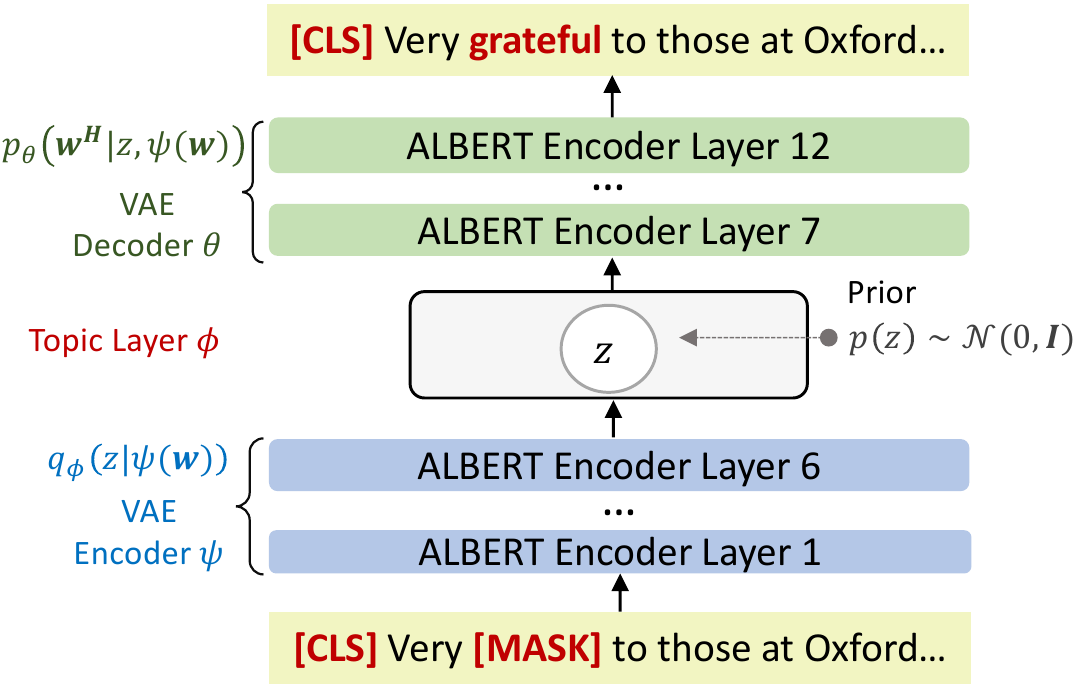}
\caption{\textsc{VADet} in masked language model learning. The latent variables are encoded via the topic layers incorporated into the masked language model. } \vspace{-10pt}
\label{fig:VADet-unsupervised} 
\end{figure}

\paragraph{\textsc{VADet} in the masked LM learning}

\begin{figure*}[tb]
\centering
\includegraphics[width=0.9\textwidth]{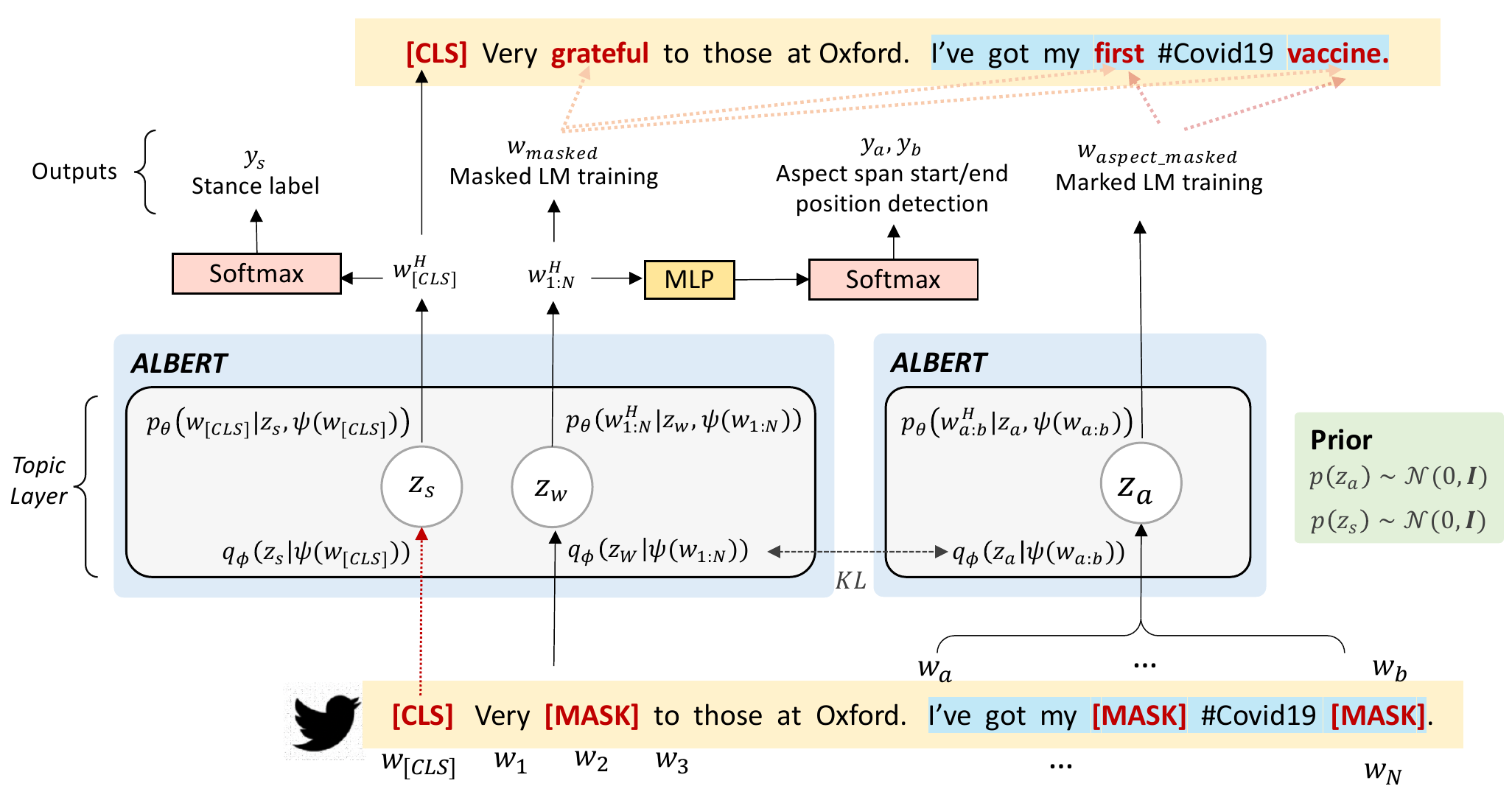}
\caption{\textsc{VADet} in supervised learning. The text segment highlighted in blue is the annotated aspect span. The right part learns latent aspect topic $z_a$ from aspect text span $[w_a:w_b]$ only under masked LM learning. The left part learns jointly latent stance topic $z_s$ and latent aspect topic $z_w$ from the whole input text, and trained simultaneously for stance classification and aspect start/end position detection.} \vspace{-10pt}
\label{fig:VADet-supervised} 
\end{figure*}

We insert a topic layer into a pre-trained language model such as ALBERT, as shown in Figure \ref{fig:VADet-unsupervised}, 
allowing the network to leverage pre-trained information while fine-tuned on an in-domain corpus. 
We assume that there is a continuous latent variable $z$ involved in the language model to reconstruct the original text from the masked tokens. 
We retain the weights of a language model and learn the latent representation during the fine-tuning. 
More concretely, the topic layer partitions a language model into lower layers and higher layers denoted as $\psi$ and $\theta$, respectively. The lower layers constitute the Encoder that parameterizes the variational posterior distribution denoted as $q_\phi(z|\psi(\textbf{w}))$, while the higher layers reconstruct the input tokens, which is referred to as the Decoder. 

The objective of VAE is to minimize the KL-divergence between the variational posterior distribution and the approximated posterior. This is equivalent to maximizing the Evidence Lower BOund (ELBO) expressed as:
\begin{equation}\small
\label{eqs:1}
\mathbb{E}_{q_\phi(z|\psi(\textbf{w}))}[\mathrm{log}\;p_\theta(\textbf{w}^H|z, \psi(\textbf{w}))] 
- \mathrm{KL}[q_\phi(z|\psi(\textbf{w}))||p(z)],
\end{equation}
where $q_\phi(z|\psi(\textbf{w}))$ is the encoder and $p_\theta(\textbf{w}^H|z, \psi(\textbf{w}))$ is the decoder. Here, $\textbf{w} = \lbrack w_{\texttt{CLS}}, w_{1:n} \rbrack$, since the special classification embedding $w_{\texttt{CLS}}$ is automatically prepended to the input sequence~\cite{devlin2019bert}, $\textbf{w}^H$ denotes the reconstructed input.

Following~\cite{kingma2014auto}, we choose a standard Gaussian distribution as the prior, denoted as $p(z)$, and the diagonal Gaussian distribution $z \sim \mathcal{N}(\mu_{\phi}(\psi(\textbf{w})), \sigma^2_{\phi}(\psi(\textbf{w})))$ as the variational distribution. 
The decoder computes the probability of the original token given the latent variable sampled from the Encoder. We use the Memory Scheme~\cite{li-etal-2020-optimus} to concatenate $z$ and $\psi(\textbf{w})$, making the latent representation compatible for higher layers of the language model. Then the latent presentation $z$ is passed to $\theta$ to reconstruct the original text. 







\paragraph{\textsc{VADet} with disentanglement of aspect and stance} \lx{One of the training objectives of vaccine attitude detection is to detect the text span that indicates the aspect and to predict the associated stance label. Existing approaches rely on structured annotations to indicate the boundary and dependency between aspect span and opinion words~\cite{xu-etal-2020-position,barnes-etal-2021-structured}, or use a two-stage pipeline to detect the aspect span and the associated opinion separately~\cite{Peng_Xu_Bing_Huang_Lu_Si_2020}. The problem is that 
the opinion expressed in a tweet and the aspect span often overlap. To mitigate this issue, we instead separate the stance and aspect from their representations in the latent semantic space, that is, disentangling latent topics learned by \textsc{VADet} into latent stance topics and latent aspect topics.} 
A recent study in disentangled representation learning~\cite{locatello2019challenging} shows that unsupervised learning of disentangled representations is theoretically impossible from i.i.d. observations without inductive biases, such as grouping information \cite{bouchacourt2018multi} or access to labels \cite{locatello2020disentangling,trauble2021disentangled}. 
As such, we extend our model to a supervised setting in which disentanglement of the latent vectors can be trained on annotated data.

Figure~\ref{fig:VADet-supervised} outlines the overall structure of \textsc{VADet} in the supervised setting. On the right hand side, we show \textsc{VADet} learned from the annotated aspect text span $[w_a:w_b]$ under masked LM learning. The latent variable $z_a$ encodes the hidden semantics of the aspect expression. We posit that the aspect span is generated from a latent representation with a standard Gaussian distribution being its prior. The ELBO for reconstructing the aspect text span is:

\begin{equation}\small
\begin{aligned}
\mathcal{L}_A = \mathbb{E}_{q_\phi(z_a|\psi(w_{a:b}))}[\mathrm{log}\;p_\theta(w_{a:b}^H|z_a, \psi(w_{a:b}))]\\
- \mathrm{KL}[q_\phi(z_a|\psi(w_{a:b}))||p(z_a)],
\end{aligned}
\end{equation}

\noindent where $w_{a:b}^H$ denotes the reconstructed aspect span. 
Ideally, the latent variable $z_a$ does not encode any stance information and only captures the aspect mentioned in the sentence. \lxCR{Therefore, the $z_s$ for the language model on the right hand side is detached and the reconstruction loss for \texttt{[CLS]} is set free.}

On the left hand side of Figure~\ref{fig:VADet-supervised}, we train \textsc{VADet} on the whole sentence. The input to \textsc{VADet} is formalized as:
`\texttt{[CLS]} \textsf{text}'. 
Instead of mapping an input to a single latent variable $z$, as in masked LM learning of \textsc{VADet}, the input is now mapped to a latent variable decomposing into two components, $[z_s, z_w]$, one for the stance and another for the aspect. 
We use a conditionally factorized Gaussian prior over the latent variable $z_w \sim p_\theta(z_w|w_{a:b})$, which enables the separation of $z_s$ and $z_w$ since the diagonal Gaussian is factorized and the conditioning variable $w_{a:b}$ is observed.


\lx{We establish an association between $z_w$ and $z_a$ by specifying $p_\theta(z_w|w_{a:b})$ to be the encoder network of $q_\phi(z_a|w_{a:b})$, since we want the latent semantics of aspect span to encourage the disentanglement of attitude in the latent space.} 
\lx{In other words,} the prior of $z_w$ is configured as the approximate posterior of $z_a$ to enforce the association between the disentangled aspect in sentence and the \textit{de facto} aspect. As a result, the ELBO for the original text is written as
\begin{equation}\small\label{eqs:2}
\begin{aligned}
\mathbb{E}_{q_\phi(z_w|\psi(\textbf{w}))}&\lbrack\mathrm{log}\;p_\theta(\mathbf{w}^H|z_w, \psi(\textbf{w}))\rbrack \\
    - &\mathrm{KL}[q_\phi(z_w|\psi(\textbf{w}))||q_\phi(z_w|\psi(w_{a:b}))],
\end{aligned}
\end{equation}
where $\mathbf{w}^H$ denotes the reconstructed input text, $z_w|\textbf{w} \sim \mathcal{N}(\mu_\phi(\psi(\textbf{w})), \sigma^2_\phi(\psi(\textbf{w})))$. The KL-divergence allows for some variability since there might be some semantic drift from the original semantics when the aspect span is placed in a longer sequence.

The annotation of the stance label provides an additional input. To exploit this \lx{inductive bias}, we enforce the constraint that $z_s$ participates in the generation of \texttt{[CLS]}, which follows an approximate posterior $q_\phi(z_s|\psi(\textbf{w}_{\texttt{[CLS]}}))$. We place the standard Gaussian as the prior over $z_s \sim \mathcal{N}(\textbf{0}, \textbf{I})$ and obtain the ELBO
{\begin{equation}\small\label{eqs:3}
\begin{aligned}
\mathbb{E}_{q_\phi(z_s|\psi(\mathbf{w}_{\texttt{[CLS]}}))}[\mathrm{log}\;p_\theta(w_{\texttt{[CLS]}}^H|z_s, \psi(\textbf{w}_{\texttt{[CLS]}}))]\\
- \mathrm{KL}[q_\phi(z_s|\psi(\mathbf{w}_{\texttt{[CLS]}}))||p(z_s)]
\end{aligned}
\end{equation}
}
Since the variational family in Eq.~\ref{eqs:1} are Gaussian distributions with diagonal covariance, the joint space of $[z_s, z_w]$ factorizes as $q_\phi(z_s,z_w|\psi(\textbf{w})) = q_\phi(z_s|\psi(\textbf{w}))q_\phi(z_w|\psi(\textbf{w}))$ \cite{nalisnick2016approximate}. Assuming $z_w$ to be solely dependent on $\psi(w_{1:n})$, we obtain the ELBO for the entire input sequence:
\begin{equation}\small
\begin{aligned}
\mathcal{L}_S = & \  \mathbb{E}_{q_\phi(z_w)}\mathbb{E}_{q_\phi(z_s)}[\mathrm{log}\;p_\theta(\mathbf{w}^H|z, \psi(\mathbf{w}))]\\
& - \mathrm{KL}\left[ q_\phi(z_w|\psi(w_{1:n}))||q_\phi(z_w|\psi(w_{a:b}))\right]\\
& - \mathrm{KL}[q_\phi(z_s|\psi(\mathbf{w}))||p(z_s)].
\end{aligned}
\end{equation}
Note that the expectation term can be decomposed into the expectation term in Eq.~\ref{eqs:2} and Eq.~\ref{eqs:3} according to the decoder structure. \lx{For the full derivation, please refer to Appendix A.}

Finally, \lx{we perform stance classification and classification for the starting and ending position over the aspect span of a tweet.} We use negative log-likelihood loss for \lx{both} the stance label and aspect span:
\begin{equation}\small
\begin{aligned}
\mathcal{L}_s = &-\mathrm{log}\;p(y_s|w^H_{\texttt{[CLS]}}), \nonumber \\
\mathcal{L}_a = &- \mathrm{log}\;p(y_a|\mathrm{MLP}(w_{1:n}^H)) - \mathrm{log}\;p(y_b|\mathrm{MLP}(w_{1:n}^H)), \nonumber
\end{aligned}
\end{equation}
where MLP is a fully-connected feed-forward network with tanh activation, $y_s$ is the predicted stance label, $y_a$ and $y_b$ are the starting and ending position of the aspect span. The \lx{overall} training objective \lx{in the supervised setting} is:
\begin{equation}\small
    \mathcal{L} = \mathcal{L}_s + \mathcal{L}_a - \mathcal{L}_S - \mathcal{L}_A\nonumber
    \vspace{-10pt}
\end{equation}





\label{sec:3d1}


\section{Experiments}

We present below the experimental setup and evaluation results.

\subsection{Experimental Setup}
\label{sec:4d1}

\begin{table}
    \centering
    \resizebox{0.4\textwidth}{!}{
\begin{tabular}{lrrrr}
\toprule
& \multicolumn{2}{c}{\multirow{1}{*}{VAD}} & \multicolumn{2}{c}{\multirow{1}{*}{VC}} \\
\cmidrule(lr){2-3} \cmidrule(lr){4-5}
Specification & \multicolumn{1}{c}{Train} & Test & Train & Test \\
\midrule
\# tweets &2000& 800 & 1162 & 531 \\
\enskip \# anti-vac. & 638 & 240 & 822 & 394\\
\enskip \# neutral & 142 & 76 & 41& 27\\
\enskip \# pro-vac. & 1220 & 484 & 299 & 110\\
\midrule
Avg. length & 33.5 & 34.13 & 29.6 & 30.24\\
\enskip len(aspect) &17.5 & 18.75 & 1.03 & 1.08\\
\enskip len(opinion) &27.97 & 29.01 & 3.25 & 3.15\\
\midrule
\# tokens & 67k & 27.3k & 34.4k& 16.8k\\
\bottomrule
\end{tabular}}
    \caption{Dataset Statistics. `\# tweets' denotes the number of tweets in VAD, and for VC it is the number of sentences. `anti-vac.' means \emph{anti-vaccination} while `pro-vac.' means \emph{pro-vaccination}. `Avg. length' and `\# token' measure the number of word tokens.}
    \label{tab:minus1}
\end{table}
\paragraph{Datasets}

We evaluate our proposed \textsc{VADet} and compare it against baselines on two vaccine attitude datasets. 

\noindent\underline{VAD} is our constructed \textbf{V}accine \textbf{A}ttitude \textbf{D}ataset. 
Following~\cite{info:doi/10.2196/26627}, we crawl tweets using the Twitter streaming API with $60$ pre-defined keywords\footnote{The full keyword list and the details of dataset construction are presented in Appendix B.} relating to COVID-19 vaccines (e.g., \textit{Pfizer}, \textit{AstraZeneca}, and \textit{Moderna}). Our final dataset comprises $1.9$ million English tweets collected between \lx{February} 7th and April 3rd, 2021. We randomly sample a subset of tweets for annotation. 
Upon an initial inspection, we found that over $97\%$ of tweets mentioned only one aspect. As such, we annotate each tweet with a stance label and a text span characterizing the aspect. In total, 2,800 tweets have been annotated in which 2,000 are used for training and the remaining 800 are used for testing. \lx{The statistics of the dataset is listed in Table~\ref{tab:minus1}. The stance labels are imbalanced. On the other hand, the average opinion length is longer than the average aspect length, and is close to the average tweet length.} For the purpose of evaluation on tweet clustering and latent topic disentanglement, we \lx{further} annotate tweets with a categorical label indicating the aspect category. Inspired by~\cite{morante-etal-2020-annotating}, we identify $24$ aspect categories\footnote{The full list of aspect categories is shown in Table \ref{tab:aspectCat}.} and each tweet is annotated with one of these categories. It is worth mentioning that aspect category labels are not used for training. 

\noindent\underline{VC}~\cite{morante-etal-2020-annotating} is a vaccination corpus consisting of 294 Internet documents about online vaccine debate annotated with events, 210 of which are annotated with opinions (in the form of text spans) towards vaccines. 
The stance label is considered to be the stance for the whole sentence. Those sentences with conflicting stance labels are regarded as neutral. We split the dataset into a ratio of 2:1 
for training and testing. This eventually left us with 1,162 sentences for training and 531 sentences for testing.

\paragraph{Baselines}
\label{ss:baselines}
We compare the experimental results with the following baselines:

\noindent\underline{BertQA} \cite{ijcai2018-583}: a pre-trained language model well-suited for span detection. \lx{With BertQA, attitude detection is performed by first classifying stance labels then predicting the answer queried by the stance label. The text span is configured as the ground-truth answer.} We rely on its HuggingFace\footnote{\url{https://huggingface.co/transformers/model_doc/albert.html\#albertforquestionanswering}}~\cite{wolf-etal-2020-transformers} implementation. We employ ALBERT~\cite{Lan2020ALBERT} as the backbone language model for both BertQA and \textsc{VADet}. 
  
\noindent\underline{ASTE} \cite{Peng_Xu_Bing_Huang_Lu_Si_2020}: a pipeline approach consisting of aspect extraction~\cite{ijcai2018-583} and sentiment labelling~\cite{li-etal-2018-transformation}.

\paragraph{Evaluation Metrics}

For stance classification, we use accuracy and Macro-averaged F1 score. For aspect span detection, we follow~\citet{rajpurkar-etal-2016-squad} in adopting exact match (EM) accuracy of the starting-ending position and Macro-averaged F1 score of the overlap between the prediction and ground truth aspect span. For tweet clustering, we follow~\citet{pmlr-v48-xieb16} and \citet{zhang-etal-2021-supporting} and use the Normalized Mutual Information (NMI) metric to measure \lx{how the clustered group aligns with ground-truth categories}. In addition, we also report the clustering accuracy.  









\subsection{Experimental Results}

In all our experiments, \textsc{VADet} is firstly pre-trained in an unsupervised way on our collected 1.9 million tweets before fine-tuned on the annotated training set from the VAD or VC corpora.

\paragraph{Stance Classification and Aspect Span Detection}
\begin{table}
\centering
\resizebox{0.4\textwidth}{!}{
\begin{tabular}{lcccc}
\toprule
\multicolumn{1}{l}{\multirow{1}{*}{Model}} & \multicolumn{2}{c}{\multirow{1}{*}{VAD}} & \multicolumn{2}{c}{\multirow{1}{*}{VC}} \\
\cmidrule(lr){2-3} \cmidrule(lr){4-5}
\textbf{\textit{Stance}} & \multicolumn{1}{c}{Acc.} & \multicolumn{1}{c}{F1} & Acc. & \multicolumn{1}{c}{F1} \\
\midrule
BertQA & 0.754& 0.742 &0.719& 0.708\\
ASTE &0.723& 0.710 &0.704& 0.686\\
\textsc{VADet} & \textbf{0.763} & \textbf{0.756} & \textbf{0.727} & \textbf{0.713} \\
\midrule
\textbf{\textit{Aspect Span}}& \multicolumn{1}{c}{Acc.} & \multicolumn{1}{c}{F1} & Acc. & \multicolumn{1}{c}{F1} \\
\midrule
BertQA & 0.546& 0.722 &0.525& 0.670\\
ASTE &0.508& 0.684 &0.467& 0.652\\
\textsc{VADet} & \textbf{0.556} & \textbf{0.745} & \textbf{0.541} & \textbf{0.697} \\
\midrule
\textbf{\textit{Cluster}} & Acc.& NMI & Acc. & NMI\\
\midrule
DEC (BertQA) & 0.633 & 58.1 & 0.586 & 52.8\\
K-means (BERT) & 0.618 & 56.4 & 0.571 & 50.1\\
DEC (\textsc{VADet}) & \textbf{0.679} & \textbf{60.7} & \textbf{0.605} & \textbf{54.7} \\
\bottomrule
\end{tabular}}
\caption{Results for stance classification, aspect span extraction and aspect clustering on both VAD and VC corpora. } \vspace{-10pt} 
\label{tab:2}
\end{table}

In Table~\ref{tab:2}, \lx{we report the performance on attitude detection.}
In stance classification, our model outperforms both baselines with more significant improvements on ASTE. 
On aspect span extraction, \textsc{VADet} yields even more noticeable improvements, with a $2.3\%$ increase in F1 over BertQA on VAD, and $2.7\%$ on VC. \lx{These results indicate that the successful prediction relies on the hidden representation learned in the unsupervised training. The disentanglement of stance and aspect may have also contributed to the improvement.}

\begin{figure}[ht]
\centering
\subfloat[VAD\textsc{et}]{\includegraphics[width=0.24\textwidth]{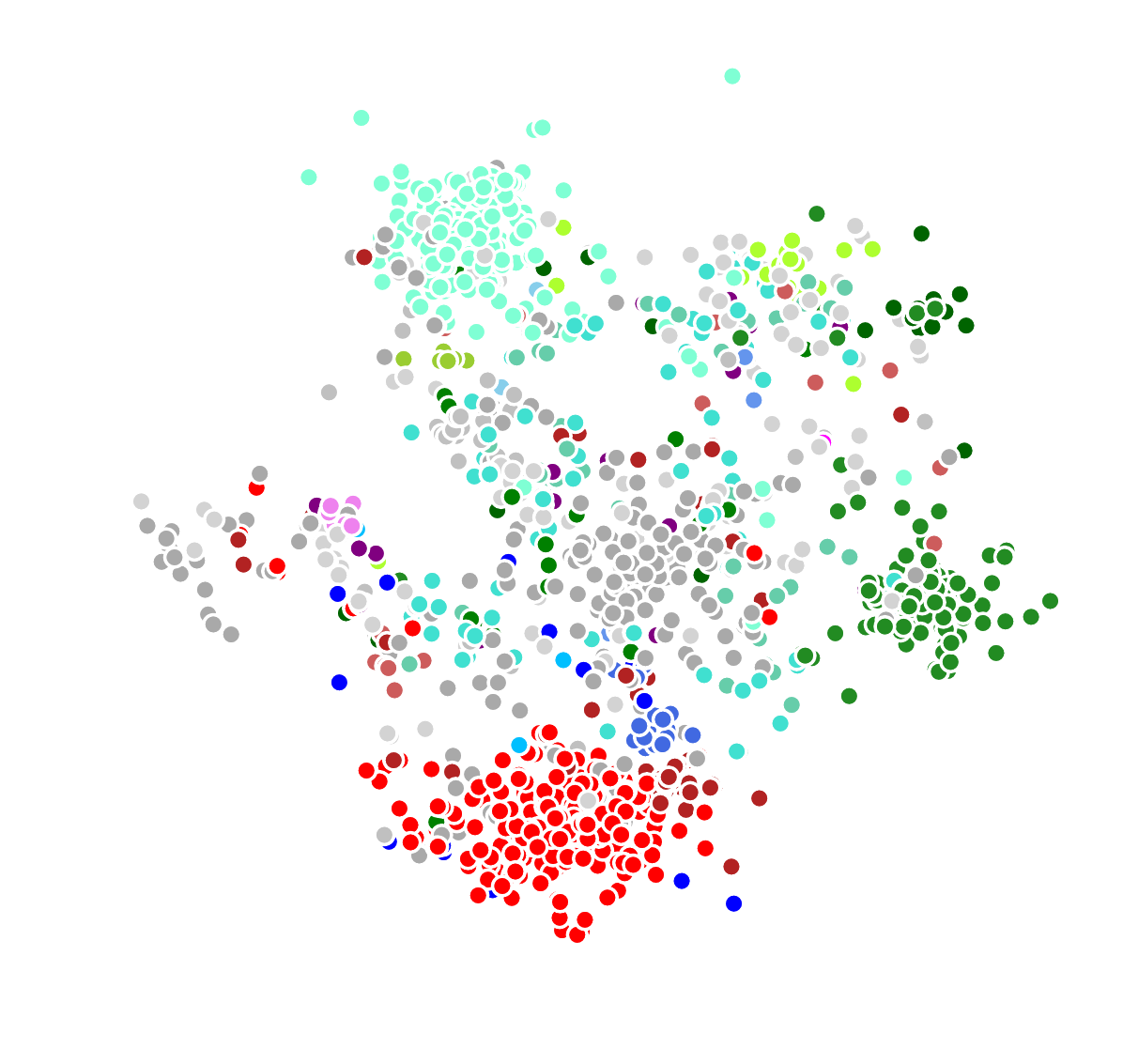}}
\subfloat[BertQA]{\includegraphics[width=0.24\textwidth]{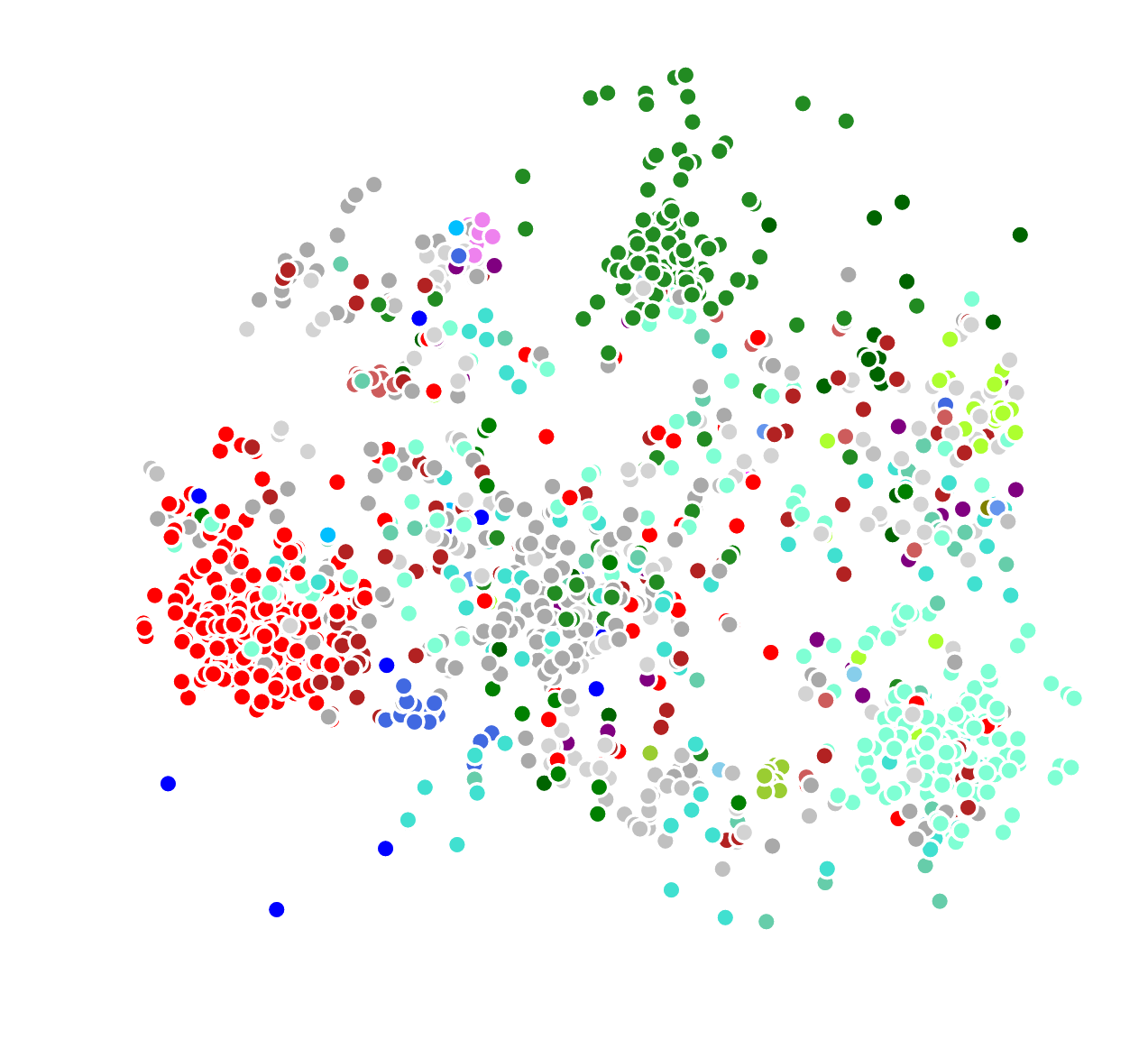}}
\caption{Clustered groups of \textsc{VADet} and BertQA \lx{on the VAD dataset}. Each
color indicates a ground truth aspect category. The clusters are dominated by: \textit{(1) Red: the (adverse) side effects of vaccines; (2) Green: explaining personal experiences with any aspect of vaccines; and (3) Cyan: the immunity level provided by vaccines}.}
\label{fig:2}
\end{figure}

\paragraph{Clustering}
To assess whether the learned latent aspect topics would allow meaningful categorization of documents into attitude clusters, we perform clustering using the disentangled representations that encode aspects, i.e., $z_w$. Deep Embedding Clustering (DEC)~\cite{pmlr-v48-xieb16} is employed as the backend. For comparison, we also run DEC on the aspect representations of documents returned by BertQA. For each document, its aspect representation is obtained by averageing over the fine-tuned ALBERT representations of the constituent words in its aspect span. To assess the quality of clusters, we need the annotated aspect categories for documents in the test set. In VAD, we use the annotated aspect labels as the ground-truth categories whereas in VC we use the annotated event types. 
Results are presented in the lower part of Table~\ref{tab:2}. We found a prominent increase in NMI score over the baselines. Using the learned latent aspect topics as features,  DEC (\textsc{VADet}) outperforms DEC (BertQA) by $4.6\%$ and $1.9\%$ in accuracy on VAD and VC, respectively. We also notice that using K-means as the clustering approach directly on the BERT-encoded tweet representations gives worse results compared to DEC. A similar trend is observed on the NMI metric. The improvements are shown visually in Figure~\ref{fig:2} where the clustered groups produced by \textsc{VADet} are more identifiable. In the absence of categorical labels, the perspective expressed by each group can be inferred from the constituent tweets. For example, the tweet \textit{`@user Georgian nurse dies of allergic reaction after receiving AstraZeneca Covid19 vaccine'} lies in the centroid of the red group, which relates to safety concerns.

\begin{figure}[ht]
\centering
\includegraphics[width=0.47\textwidth]{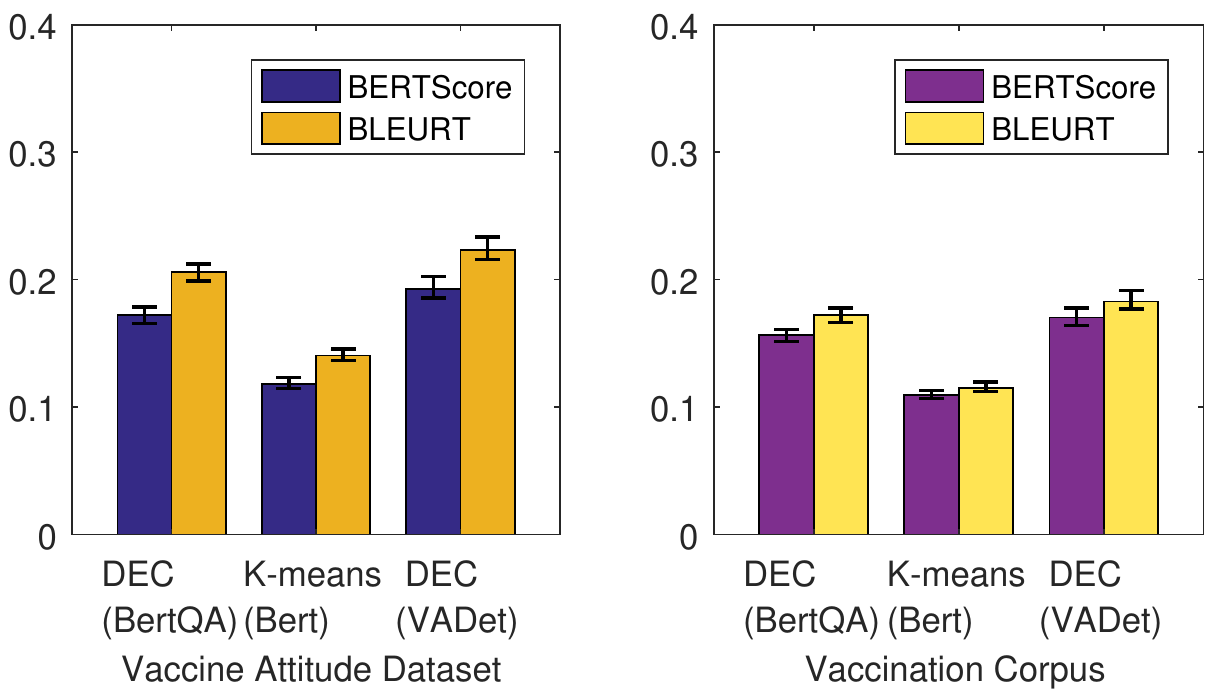}
\caption{Semantic coherence evaluated in two metrics.}
\label{fig:5}
\end{figure}

\paragraph{Cluster Semantic Coherence Evaluation}
The semantic coherence is the extent to which tweets within a cluster belong to each other, which is employed as an evaluation metric for cluster quality evaluation in an unsupervised way. Recent work of \citet{bilal-etal-2021-evaluation} found that Text Generation Metrics (TGMs) align well with human judgement in evaluating clusters in the context of microblog posts. TGM by definition measures the similarity between the ground-truth and the generated text. The rationale is that a high TGM score means sentence pairs are semantically similar. 
Here, two metrics are used: \textit{BERTScore}, which calculates the similarity of two sentences as a sum of cosine similarities between their tokens’ embeddings~\cite{DBLP:conf/iclr/ZhangKWWA20}, and \textit{BLEURT}, a pre-trained adjudicator that fine-tunes BERT on an external dataset of human ratings~\cite{sellam-etal-2020-bleurt}. As in~\cite{bilal-etal-2021-evaluation}, we adopt the Exhaustive Approach that for a cluster $C$, its coherence score is the average TGM score of every possible tweet pair in the cluster: 
\begin{equation*}
f(C) = \frac{1}{N^2}\sum_{i,j\in [1, N], i<j}\mathrm{TGM}(\mathrm{tweet}_i, \mathrm{tweet}_j).
\end{equation*}
Figure~\ref{fig:5} shows the BERTScore and the BLEURT score of VAD\textsc{et} and baselines on two datasets. The VAD\textsc{et} shows consistent improvements across the datasets. This indicates that tweets clustered using the latent aspect topics generated by VAD\textsc{et} are semantically more similar, thus validating the assumption that disentangled representations are more effective in bringing together tweets of a similar gist. 

\paragraph{Conditional Perplexity}
Few metrics have been proposed to evaluate the quality of disentangled representations~\cite{pergola-etal-2021-disentangled}. Therefore, we adopt the language model perplexity conditioned on $z_a$ to evaluate the extent to which the disentangled representation improves language generation on held-out data. Perplexity is widely used in the literature of text style transfer~\cite{john-etal-2019-disentangled,ijcai2020-526}, where the probability of the generated language is calculated conditioned on the controlled latent code. A lower perplexity score indicates better language generation performance. Following \citet{john-etal-2019-disentangled}, we compute an estimated aspect vector $\hat{z}_a^{(k)}$ of a cluster $k$ in the training set as
\begin{equation*}
    \hat{z}_a^{(k)} = \frac{\sum_{i\in \mathrm{cluster}\;k} z_{a,i}^{(k)}}{\mathrm{\mypound\;tweets\;in\;cluster}\;k},
\end{equation*}
where $z_{a,i}^{(k)}$ is the learned aspect vector of the $i$-th tweet in the $k$-th cluster. For the stance vector $z_s$, we sample one value per tweet. 
The stance vector is concatenated with the aspect vector $\hat{z}_a^{(k)}$ to calculate the probability of generating the held-out data, i.e., the testing set. For the baseline models, we choose $\beta$-VAE~\cite{DBLP:conf/iclr/HigginsMPBGBML17} and SCHOLAR~\cite{card-etal-2018-neural}. We train $\beta$-VAE on the same data with $\beta$ set to different values. SCHOLAR is trained on tweet content and stance labels. For both the baselines we use ELBO on the held-out data as an upper bound on perplexity.

\begin{figure}[ht]
\centering
\includegraphics[width=0.48\textwidth]{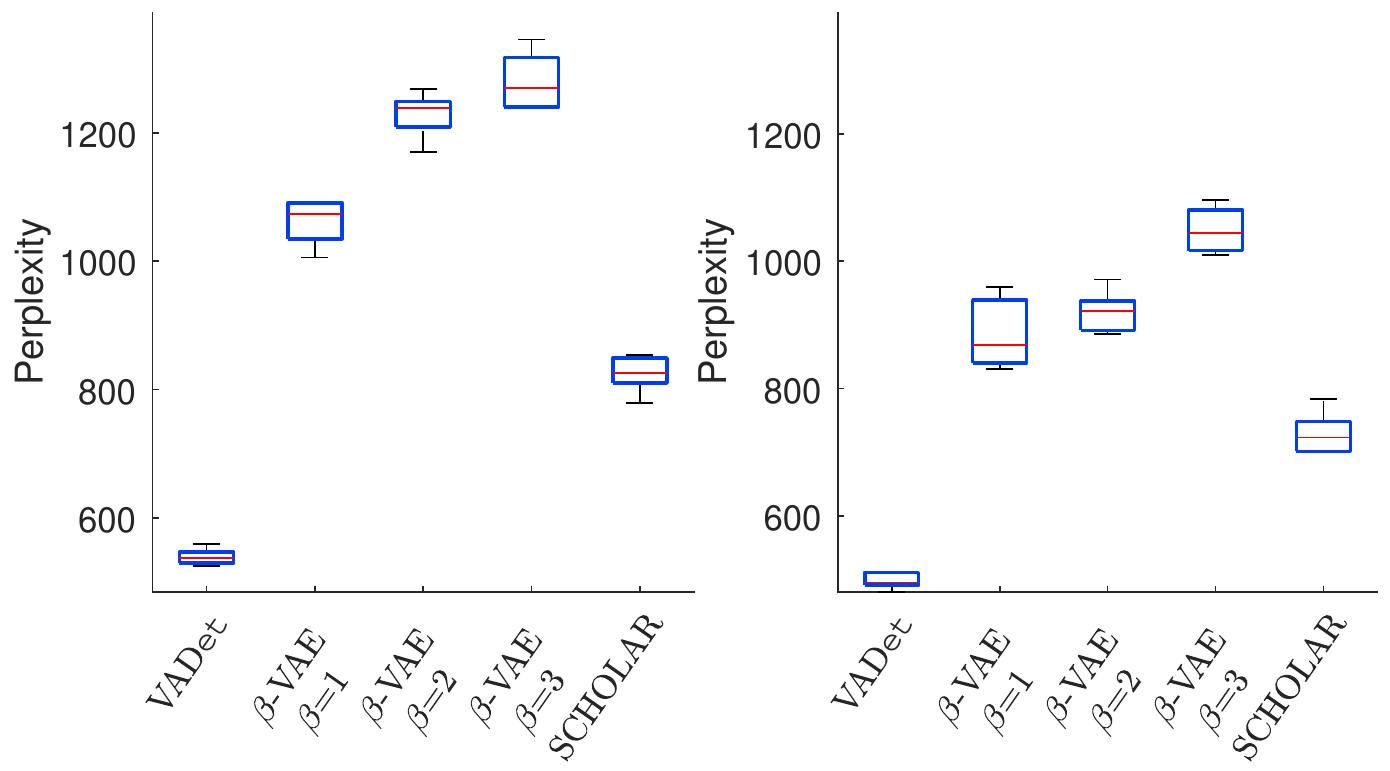}
\caption{Conditional perplexity on two corpora.}
\label{fig:6}
\end{figure}

Figure~\ref{fig:6} plots the perplexity score achieved by all the methods. 
Our model achieves the lowest perplexity score on both datasets. It managed to decrease the perplexity value by roughly 200 compared to the baseline models. SCHOLAR outperforms $\beta$-VAE under three settings of $\beta$ value. We speculate that this might be due to the incorporation of the class labels in the training of SCHOLAR. Nevertheless, VAD\textsc{et} produces congenial sentences in aspect groups, with latent codes tweaked to proxy centroids, showing that the disentangled representation does capture the desired factor.

\paragraph{Ablations}
We conduct ablation studies to investigate the effect of semi-supervised learning that uses the variational latent representation learning approach and aspect-stance disentanglement on the latent semantics. We study their effects on stance classification and aspect span detection. The results are reported in Table~\ref{tab:3}. 



\begin{table}[h!]
\centering
\centering
\resizebox{0.45\textwidth}{!}{
\begin{tabular}{lcccc}
\toprule
Model  & \multicolumn{2}{c}{\multirow{1}{*}{VAD}} & \multicolumn{2}{c}{\multirow{1}{*}{VC}} \\
\cmidrule(lr){2-3} \cmidrule(lr){4-5} 
\textbf{\emph{Stance}} & \multicolumn{1}{c}{Acc.} & \multicolumn{1}{c}{F1} & Acc. & \multicolumn{1}{c}{F1} \\
\midrule
VAD\textsc{et} & \textbf{0.763}& \textbf{0.756} &0.727& 0.713\\
VAD\textsc{et}-D & 0.751& 0.746 &\textbf{0.736}& \textbf{0.716}\\
VAD\textsc{et}-U &0.741& 0.734 &0.712& 0.698\\
\midrule
\textbf{\emph{Aspect Span}} & \multicolumn{1}{c}{Acc.} & \multicolumn{1}{c}{F1} & Acc. & \multicolumn{1}{c}{F1} \\
\midrule
VAD\textsc{et} & \textbf{0.556}& \textbf{0.745} & \textbf{0.541} & \textbf{0.697} \\
VAD\textsc{et}-D & 0.540& 0.728 &0.537& 0.684 \\
VAD\textsc{et}-U &0.528& 0.712 &0.525& 0.653 \\
\bottomrule
\end{tabular}}
\caption{Results of stance classification and aspect span detection of VAD\textsc{et} without disentanglement (-D) or unsupervised pre-training (-U). } 
\label{tab:3}
\end{table}

We can observe that on VAD without disentangled learning or unsupervised pre-training results in the degradation of the stance classification performance. However, on VC, we see a slight increase in classification accuracy without disentangled learning. We attribute this to the vagueness of the stance which might cause the model to disentangle more than it should be. On the aspect span detection task, we observe consistent performance drop across all metrics and on both datasets. In particular, without the pre-training module, the performance drops more significantly. 
These results indicate that semi-supervised learning is highly effective with VAE, and the disentanglement of stance and aspect serves as a useful component, which leads to noticeable improvements.







\section{Conclusions}
In this work, we presented a semi-supervised model to detect user attitudes and distinguish aspects of interest about vaccines on social media. We employed a Variational Auto-Encoder to encode the main topical information into the language model by unsupervised training on a massive, unannotated dataset. 
The model is then further trained under a semi-supervised setting that leverages annotated stance labels and aspect spans to induce the disentanglement between stances and aspects in a latent semantic space. We empirically showed the benefits of such an approach for attitude detection and aspect clustering over two vaccine corpora. Ablation studies show that disentangled learning and unsupervised pre-training are important to effective vaccine attitude detection. 
Further investigations on the quality of the disentangled representations verify the effectiveness of the disentangled factors. 
While our current work mainly focuses on short text of social media data where a sentence is assumed to discuss a single aspect, it would be interesting to extend our model to deal with longer text such as online debates in which multiple arguments or aspects may appear in a single sentence. 

\section*{Acknowledgements}
 \lxCR{This work was funded by the the UK Engineering and Physical Sciences Research Council (grant no. EP/T017112/1, EP/V048597/1). LZ is supported by a Chancellor's International Scholarship at the University of Warwick. YH is supported by a Turing AI Fellowship funded by the UK Research and Innovation (grant no. EP/V020579/1). } 















\bibliography{anthology,custom}
\bibliographystyle{acl_natbib}

\clearpage
\appendix

\setcounter{table}{0}
\renewcommand{\thetable}{A\arabic{table}}

\section{Derivation of the Decomposed ELBO}
\label{sec:appendix}

Unsupervised training is based on maximizing the Evidence Lower Bound (ELBO):
\begin{equation}
\begin{aligned}
\mathbb{E}_{q_\phi(z_s, z_w|\psi(\textbf{w}))}[\mathrm{log}\;p_\theta(\textbf{w}|z_s,z_w, \psi(\textbf{w}))] \\
 - \mathrm{KL}[q_\phi(z_s,z_w|\psi(\textbf{w}))||p(z_s, z_w)], \nonumber
\end{aligned}
\end{equation}
where $z$ is partitioned into $z_s$ and $z_w$. Like standard VAE~\cite{kingma2014auto}, the variational distribution is a multivariate Gaussian with a diagonal covariance:
\begin{equation}
q_\phi(z_s, z_w|\psi(\textbf{w})) = \mathcal{N}(z_s, z_w|\mu, \sigma^2 I), \nonumber
\end{equation}
where $\mu = [\mu^s, \mu^w]$ and $\sigma = [\sigma^s, \sigma^w]$. Since the coveriance matrix is diagonal, 
$z_s$ and $z_w$ are uncorrelated. Therefore, the joint probability is decomposed into:
\begin{equation}
    q_\phi(z_s, z_w|\psi(\textbf{w})) = q_\phi(z_s|\psi(\textbf{w}))q_\phi(z_w|\psi(\textbf{w})), \nonumber
\end{equation}
where $q_\phi(z_s|\psi(\textbf{w}))=\mathcal{N}(z_s|\mu^s, {\sigma^s})$, $\phi$ are the variational parameters. The prior of $[z_s, z_w] \sim \mathcal{N}(z_s, z_w| \mathbf{0}, I)$ can also be decomposed into the product of $p(z_s)$ and $p(z_w)$, then the KL term becomes:
\begin{equation}
\small
\begin{aligned}
    \mathrm{KL}[q_\phi(z_s|\psi(\textbf{w}))||p(z_s)] + \mathrm{KL}[q_\phi(z_w|\psi(\textbf{w}))||p(z_w)].
\end{aligned}\nonumber
\end{equation}
As for the decoder $p_\theta(\textbf{w}|z_s,z_w, \psi(\textbf{w}))$, the reconstruction of each masked token and $w_{\texttt{[CLS]}}$ are independent from each other, i.e., they are not predicted in an autoregressive way. Therefore, the joint probability is decomposed into:
\begin{equation}
\begin{aligned}
& p_\theta(\textbf{w}|z_s,z_w, \psi(\textbf{w}))\\
& = p_\theta(w_{\texttt{[CLS]}}|z_s,z_w, \psi(\textbf{w}))\, p_\theta(w_{1:n}|z_s,z_w, \psi(\textbf{w}))
\end{aligned}\nonumber
\end{equation}
We customize the decoder network to make $w_{\texttt{[CLS]}}$ solely dependent on $z_s$, and obtain
\begin{equation}
\begin{aligned}
\mathbb{E}_{q_\phi(z_s)} \mathbb{E}_{q_\phi(z_w)}[\mathrm{log}\;p_\theta(w_{\texttt{[CLS]}}|z_s, \psi(\textbf{w})) \ + \\
\mathrm{log}\;p_\theta(w_{1:n}|z_w, \psi(\textbf{w}))]
\end{aligned}\nonumber
\end{equation}
Here, we omit $\psi(\textbf{w})$ for notational simplicity. Given the supervision of annotated aspect spans, the prior of $z_w$ is constrained by $q_\phi(z_w|\psi(w_{a:b}))$ (a.k.a., the encoder of $w_{a:b}$), this will change the KL term into:
\begin{equation}
\small
\begin{aligned}
    &\mathrm{KL}[q_\phi(z_s|\psi(\textbf{w}))||p(z_s)] \\ 
    & + \mathrm{KL}[q_\phi(z_w|\psi(w_{1:n}))||q_\phi(z_w|\psi(w_{a:b}))],
\end{aligned}\nonumber
\end{equation}
and finally the ELBO is expressed as
\begin{equation}
\begin{aligned}
& \mathbb{E}_{q_\phi(z_s)}[\mathrm{log}\;p_\theta(w_{\texttt{[CLS]}}|z_s,\psi(\textbf{w}))] \\
& + \mathbb{E}_{q_\phi(z_w)}[\mathrm{log}\;p_\theta(w_{1:n}|z_w, \psi(\textbf{w}))] \\
& - \mathrm{KL}[q_\phi(z_s|\psi(\textbf{w}))||p(z_s)]\\
& - \mathrm{KL}[q_\phi(z_w|\psi(w_{1:n}))||q_\phi(z_w|\psi(w_{a:b}))].
\end{aligned}\nonumber
\end{equation}

\section{Data Collection and Preprocessing}
We are qualified Twitter Academic Research API~\footnote{\url{https://developer.twitter.com/en/products/twitter-api/academic-research/application-info}} users. We obtained the ethical approval for our proposed research from the university's ethics committee before the start of our work. We collected tweets between February 7th and April 3rd, 2022 using 60 vaccine-related keywords. The exhaustive list is: \textit{`covid-19 vax'},
\textit{`covid-19 vaccine'},
\textit{`covid-19 vaccines'},
\textit{`covid-19 vaccination'},
\textit{`covid-19 vaccinations'},
\textit{`covid-19 jab'},
\textit{`covid-19 jabs'},
\textit{`covid19 vax'},
\textit{`covid19 vaccine'},
\textit{`covid19 vaccines'},
\textit{`covid19 vaccination'},
\textit{`covid19 vaccinations'},
\textit{`covid19 jab'},
\textit{`covid19 jabs'},
\textit{`covid vax'},
\textit{`covid vaccine'},
\textit{`covid vaccines'},
\textit{`covid vaccination'},
\textit{`covid vaccinations'},
\textit{`covid jab'},
\textit{`covid jabs'},
\textit{`coronavirus vax'},
\textit{`coronavirus vaccine'},
\textit{`coronavirus vaccines'},
\textit{`coronavirus vaccination'},
\textit{`coronavirus vaccinations'},
\textit{`coronavirus jab'},
\textit{`coronavirus jabs'},
\textit{`Pfizer vaccine'},
\textit{`BioNTech vaccine'},
\textit{`Oxford vaccine'},
\textit{`AstraZeneca vaccine'},
\textit{`Moderna vaccine'},
\textit{`Sputnik vaccine'},
\textit{`Sinovac vaccine'},
\textit{`Sinopharm vaccine'},
\textit{`Pfizer jab'},
\textit{`BioNTech jab'},
\textit{`Oxford jab'},
\textit{`AstraZeneca jab'},
\textit{`Moderna jab'},
\textit{`Sputnik jab'},
\textit{`Sinovac jab'},
\textit{`Sinopharm jab'},
\textit{`Pfizer vax'},
\textit{`BioNTech vax'},
\textit{`Oxford vax'},
\textit{`AstraZeneca vax'},
\textit{`Moderna vax'},
\textit{`Sputnik vax'},
\textit{`Sinovac vax'},
\textit{`Sinopharm vax'},
\textit{`Pfizer vaccinate'},
\textit{`BioNTech vaccinate'},
\textit{`Oxford vaccinate'},
\textit{`AstraZeneca vaccinate'},
\textit{`Moderna vaccinate'},
\textit{`Sputnik vaccinate'},
\textit{`Sinovac vaccinate'},
\textit{`Sinopharm vaccinate'}.

Only tweets in English were collected. Retweets were discarded. For pre-processing, hyperlinks, usernames and irregular symbols were removed. Emojis and emoticons were converted to their literal meanings using an emoticon dictionary\footnote{\url{https://wprock.fr/en/t/kaomoji/}}.

\section{Hyper-parameters and Training Details}
The dimensions of $z_a$, $z_w$ and $z_s$ are $768$, $768$ and $32$, respectively. For each tweet, the number of samples from $\epsilon\sim\mathcal{N}(\mathbf{0}, \mathbf{I})$ is $1$. We modified the LM-fine-tuning script\footnote{\url{https://github.com/huggingface/transformers/blob/master/examples/pytorch/language-modeling/run\_mlm.py}} from the HuggingFace library to implement VAD\textsc{et} in the masked LM learning. We use default settings for the training script (i.e., Trainer in the HuggingFace library\footnote{\url{https://huggingface.co/docs/transformers/master/en/main\_classes/trainer\#transformers.Trainer}}), except for the batch size which is set to $128$. The data pre-processor for the masked language model is the data collator for language modeling\footnote{\url{https://huggingface.co/docs/transformers/main\_classes/data\_collator}}, which provides the function of randomly masking the tokens. The tokenizer for the data collator is the ready-to-use ALBERT tokenizer\footnote{\url{https://huggingface.co/docs/transformers/master/en/model\_doc/albert\#transformers.AlbertTokenizer}}. For the pre-trained language model (i.e., ALBERT) employed in this model, we inherit the default setting from the \texttt{AlbertConfig} class. We train VAD\textsc{et} for $5$ epochs on the un-annotated corpus.

\lx{In the supervised training of VAD\textsc{et}, we use a batch size of $64$. The learning rate is initialized to $2e-5$ with a linear warm-up schedule. We employ $5$-fold training in which the training set is split into $5$ subsets, of which 4 are used for training and the rest is for validation at the end of each epoch, and the final prediction is an ensemble of 5 independently-saved models. We train each model for 5 epochs, which takes roughly $2$ hours on a node of single Nvidia RTX 2080 GPU.}

\section{Annotation Guidelines}
\lx{We invited two annotators who are PhD students and proficient in English to label each tweet with a stance label and an aspect span. Each annotator was instructed to answer four questions in a row. The four questions are:}
\begin{itemize}
    \item What is the stance towards vaccination?
    \item What is the Aspect Span? (i.e., Events or targets, it can be nouns, noun phrase, clause or sentence with verbal predicates).
    \item What is the opinion term/span? It should be opinion expressions, comprising both explicit and implicit expressions of stance.
    \item What is the Aspect category? It should be one of the pre-defined aspect categories (shown in Table~\ref{tab:aspectCat}).
\end{itemize}
The annotators have the choice to skip some of the questions if they find it difficult to answer. Taking the tweet `\textit{Very  grateful to  those  at Oxford. I’ve  got  my  first \#Covid19  vaccine.}' as an example, the annotators are expected to answer with: `\textit{Pro-vaccine}', `\textit{I’ve  got  my  first \#Covid19  vaccine}', `\textit{Very  grateful to  those  at Oxford. I’ve  got  my  first \#Covid19  vaccine}', `\textit{2}'. If an annotator chooses to skip a tweet at any step of the process, this tweet will be recorded as skipped and the annotator will not be assigned with similar tweets. 

We first had a trial run where each annotator was asked to annotate the same set of tweets. Any disagreement was recorded and discussed to refine our annotation guideline in order to achieve consistency between the annotators.


\section{Predefined Aspect Categories}

Table \ref{tab:aspectCat} shows our pre-defined aspect categories, partly inspired by~\cite{morante-etal-2020-annotating}. These categories are only used in the evaluation of tweet clustering results, not for training.

\begin{table*}[ht]
\begin{tabular}{c|l}
\multicolumn{1}{c}{\multirow{1}{*}{Label}} & \multicolumn{1}{c}{\multirow{1}{*}{Definition}} \\
\thickhline
1 &  \tabincell{l}{AstraZeneca: How health organisations/institution, communities, groups, individuals and \\ other entities position themselves towards vaccines}\\
\hline
2 & \tabincell{l}{AstraZeneca: Explaining personal experiences with any aspect of vaccines}\\
\hline
3 &  \tabincell{l}{AstraZeneca: The achievement that vaccines have brought (vaccines save lives, protect the \\community, protect future generations)}\\
\hline
4 & \tabincell{l}{AstraZeneca: The (adverse) side effects of vaccines: illnesses, symptoms, deaths}\\
\hline
5 & \tabincell{l}{AstraZeneca: The immunity level provided by vaccines}\\
\hline
6 & \tabincell{l}{AstraZeneca: The economic effect of vaccination (less illnesses, less expenses for family\\ and society)}\\
\hline
7 & \tabincell{l}{AstraZeneca: Discussing the personal freedom to choose in relation to vaccines}\\
\hline
8 & \tabincell{l}{AstraZeneca: Discussing the relation between vaccines and religion, conspiracy or moral\\ attitudes}\\
\hline
9 & \tabincell{l}{Pfizer or Moderna: How health organisations/institution, communities, groups, individuals\\ and  other entities position themselves towards vaccines}\\
\hline
10 & \tabincell{l}{Pfizer or Moderna: Explaining personal experiences with any aspect of vaccines}\\
\hline
11 & \tabincell{l}{Pfizer or Moderna: The achievement that vaccines have brought (vaccines save lives,\\ protect the community, protect future generations)}\\
\hline
12 & \tabincell{l}{Pfizer or Moderna: The (adverse) side effects of vaccines: illnesses, symptoms, deaths}\\
\hline
13 & \tabincell{l}{Pfizer or Moderna: The immunity level provided by vaccines}\\
\hline
14 & \tabincell{l}{Pfizer or Moderna: The economic effect of vaccination (less illnesses, less expenses for\\ family and society)}\\
\hline
15 & \tabincell{l}{Pfizer or Moderna: Discussing the personal freedom to choose in relation to vaccines}\\
\hline
16 & \tabincell{l}{Pfizer or Moderna: Discussing the relation between vaccines and religion, conspiracy or\\ moral attitudes}\\
\hline
17 & \tabincell{l}{Other Brands or not mentioned: How health organisations/institution, communities, \\groups, individuals and other entities position themselves towards vaccines}\\
\hline
18 & \tabincell{l}{Other Brands or not mentioned: Explaining personal experiences with any aspect of\\ vaccines}\\
\hline
19 & \tabincell{l}{Other Brands or not mentioned: The achievement that vaccines have brought (vaccines\\ save lives, protect the community, protect future generations)}\\
\hline
20 & \tabincell{l}{Other Brands or not mentioned: The (adverse) side effects of vaccines: illnesses,\\ symptoms, deaths}\\
\hline
21 & \tabincell{l}{Other Brands or not mentioned: The immunity level provided by vaccines}\\
\hline
22 &  \tabincell{l}{Other Brands or not mentioned: The economic effect of vaccination (less illnesses, less\\ expenses for family and society)}\\
\hline
23 & \tabincell{l}{Other Brands or not mentioned: Discussing the personal freedom to choose in relation\\ to vaccines}\\
\hline
24 & \tabincell{l}{Other Brands or not mentioned: Discussing the relation between vaccines and religion,\\ conspiracy or moral attitudes}\\
\thickhline
\end{tabular}
\caption{The predefined aspect categories and their definitions.}
\label{tab:aspectCat}
\end{table*}

\end{document}